\documentclass[sigconf]{acmart}
\usepackage{balance}
\usepackage{booktabs}
\usepackage{multirow}
\usepackage{siunitx}
\usepackage{xcolor}
\usepackage{makecell}
\usepackage[normalem]{ulem}

\AtBeginDocument{%
  }

\begin{document}

\title{Efficient Multi-Cohort Inference for Long-Term Effects and Lifetime Value in A/B Testing with User Learning}

\author{Dario Simionato}
% \orcid{0000-0001-5533-425X}
\authornote{Corresponding author.}
\email{dario.simionato1@h-partners.com}
\affiliation{%
  \institution{Huawei Ireland Research Centre}
  \city{Dublin}
  \country{Ireland}
}

\author{Andrea Tonon}
\email{andrea.tonon1@huawei-partners.com}
\affiliation{%
  \institution{Huawei Ireland Research Centre}
  \city{Dublin}
  \country{Ireland}
}

\author{Mingxue Wang}
\email{wangmingxue1@huawei.com}
% \authornotemark[1]
\affiliation{%
  \institution{Huawei Ireland Research Centre}
  \city{Dublin}
  \country{Ireland}
}

\author{Weiguo Wang}
\email{tom.wangweiguo@huawei.com}
\affiliation{%
  \institution{Huawei Nanjing R\&D Center}
  \city{Nanjing}
  \country{China}
}

\author{Tong Gui}
\email{guitong@huawei.com}
\affiliation{%
  \institution{Huawei Nanjing R\&D Center}
  \city{Nanjing}
  \country{China}
}

\author{Xiaoyue Li}
\email{lixiaoyue4@huawei.com}
\affiliation{%
  \institution{Huawei Nanjing R\&D Center}
  \city{Nanjing}
  \country{China}
}

\renewcommand{\shortauthors}{Dario Simionato et al.} 

\begin{abstract}
In streaming platforms churn is extremely costly, yet A/B tests are typically evaluated using outcomes observed within a limited experimental horizon. 
Even when both short- and predicted long-term engagement metrics are considered, they may fail to capture how a treatment affects users’ retention. 
Consequently, an intervention may appear beneficial in the short term and neutral in the long term while still generating lower total value than the control due to users churn.

To address this limitation, we introduce a method that estimates long-term treatment effects (LTE) and residual lifetime value change ($\Delta ERLV$) in short multi-cohort A/B tests under user learning.
To estimate time-varying treatment effects efficiently, we introduce an inverse-variance weighted estimator that combines estimates from multiple cohorts, reducing variance relative to standard approaches in the literature.
The estimated treatment trajectory is then modeled as a parametric decay to recover both the asymptotic treatment effect and the cumulative value generated over time.

Our framework enables simultaneous evaluation of steady-state impact and residual user value within a single experiment.
Empirical results show improved precision in estimating LTE and $\Delta ERLV$ and identify scenarios in which relying on either short-term or long-term metrics alone would lead to incorrect product decisions.
\end{abstract}

%%
%% The code below is generated by the tool at http://dl.acm.org/ccs.cfm.
%%
% \begin{CCSXML}
% <ccs2012>
%    <concept>
%        <concept_id>10002951.10003317.10003359.10003362</concept_id>
%        <concept_desc>Information systems~Retrieval effectiveness</concept_desc>
%        <concept_significance>500</concept_significance>
%        </concept>
%    <concept>
%        <concept_id>10002951.10003227.10003241</concept_id>
%        <concept_desc>Information systems~Decision support systems</concept_desc>
%        <concept_significance>500</concept_significance>
%        </concept>
%  </ccs2012>
% \end{CCSXML}

% \ccsdesc[500]{Information systems~Retrieval effectiveness}
% \ccsdesc[500]{Information systems~Decision support systems}

\keywords{A/B testing, user learning, long term effect, lifetime value}

% \received{20 February 2007}
% \received[revised]{12 March 2009}
% \received[accepted]{5 June 2009}

\maketitle
\begin{center}
\textit{This is the author's accepted manuscript of a paper accepted for publication in the Proceedings of the 49th International ACM SIGIR Conference on Research and Development in Information Retrieval (SIGIR 2026). The final authenticated version will be available in the ACM Digital Library.}
\end{center}

\section{Introduction}
A/B testing is a cornerstone of modern online experimentation, allowing product teams to measure the impact of product changes on user behavior in a controlled setting.
In practice, however, several real-world complexities can make interpretation challenging.
First, treatment effects often evolve over time, creating a potential mismatch between short-term results and long-term outcomes, with the latter usually being more relevant for business decisions.
Second, metrics averaged only over active users in the long term may not correctly capture the total cumulative impact of a treatment, especially when user participation changes across cohorts.

The first issue motivated researchers to focus on estimating the long-term treatment effect (LTE) from short-duration experiments where user behavior is significantly influenced by novelty or primacy effects, named as user learning \cite{kohavi2020trustworthy, dmitriev2017dirty}.
The cookie-cookie day (CCD) method~\cite{hohnhold2015CCD} was the first experimental design specifically proposed to infer long-term metrics using cohort comparisons. 
More recently, \cite{sadeghi2022Ms} showed that LTE can be estimated in the presence of user learning without requiring ad-hoc experimental designs. 
Their key idea is to analyze users w.r.t. their entry time into the experiment, rather than calendar time, and to apply a difference-in-differences (DiD) estimator across treatment and control cohorts to recover the evolution of the treatment effect.

Despite this important advancement, existing approaches remain limited in two aspects. 
First, they make inefficient use of the available data, leading to high-variance estimations of the treatment effect over time.
Second, and more fundamentally, they focus exclusively on the LTE ignoring both the impact of the treatment on user traffic and the cumulative value generated during the learning phase (i.e. when user behavior is temporarily changing).

From a business perspective, however, cumulative impact is often as important as steady-state impact. 
Lifetime value (LTV) frameworks address this objective by measuring the total value generated over time considering also user traffic rather than only asymptotic behaviors. 
Yet LTV estimation typically requires long historical horizons or detailed user-level modeling, making it poorly suited to short-term online experiments \cite{gupta2006modeling,kohavi2020trustworthy,fader2007project}.

This paper focuses on an industrial scenario for streaming services, in which a company wants to introduce an advertisement in between each stream.
While such increase may initially boost short-term ad clicks as users are not yet accustomed to such feature and are more likely to notice the ad, over time they may develop ad blindness, ignoring the ads entirely~\cite{hohnhold2015CCD, benway1998banner, cho2004people}. 
Simultaneously, repeated exposure can lead to advertisement fatigue, frustrating users and increasing churn. 
Consequently as shown in Figure~\ref{fig:metrics_comparison}, although clicks number per active user may increase in the short term and may not change in the long term, accounting for user counts and retention can reveal a decline in total engagement and lifetime value. 
This highlights the need for evaluation methods that consider both long-term metric behavior and the treatment’s effect on user traffic (i.e. how the treatment affects the user churn).

\begin{figure}[t]
    \centering
    \includegraphics[width=\linewidth]{paper_img.png}
    \caption{
        Comparison of the number of clicks $m$ observed on active users (top) and number of clicks weighted by user count (bottom). 
        In this A/B test, the treatment introduces advertisements in between streams. 
        While the computation of $m$ among active users (top) shows a positive short-term effect with no long-term downsides ($LTE \approx 0$), this view is biased by ignoring how the treatment affects user counts. 
        The second plot (bottom) accounts for user retention, revealing that the treatment significantly reduces the number of active users over time. 
        This leads to a negative overall business impact, as shown by the negative change in Expected Remaining Lifetime Value ($\Delta ERLV < 0$).
    }
    \Description{
        Two line plots comparing an A/B test treatment. The top plot shows average clicks per active user over time, where the treatment appears to improve performance with no long-term effect. The bottom plot shows total clicks adjusted by the number of users, revealing that the treatment reduces user retention over time and leads to an overall negative impact.
    }
    \label{fig:metrics_comparison}
\end{figure}

We address this problem in two ways.
First, we propose a unified framework that jointly estimates the LTE and the incremental expected residual lifetime value ($\Delta ERLV$) of users in multi-cohort A/B tests under user learning. 
Second, we introduce an inverse-variance weighted multi-cohort estimator that combines multiple unbiased estimates of the treatment evolution to improve estimation accuracy. 
Differently from CCD or DiD approaches, our multi-cohort method fully exploits all staggered cohorts reducing the variance of the estimates. 
By fitting a parametric decay model to the estimated trajectory, we recover both the asymptotic treatment effect and the cumulative value generated over time.

By simultaneously addressing the measurement objective (both LTE and $\Delta ERLV$) and the estimation procedure, the proposed method provides a more complete and reliable evaluation of A/B tests in the presence of user learning, capturing both transient dynamics and steady-state effects of product changes accounting for user churns.

\section{Problem setup}\label{sec:setup}
We consider a randomized A/B test measuring a metric $m$ (in our case, the number of ad clicks) and in which users enter the experiment over time. 
As a consequence, we analyze outcomes using a cohort framework rather than a single fixed observation window.
Let $T$ be the total duration of the A/B test, users are divided into $2T$ cohorts according to their entry time in the experiment $t_0\ (0 \le t_0 < T)$ and whether they receive their treatment (written as $T_u =1$) or they belong to the control group ($T_u=0$) throughout all the experiment. 
In particular, we define $C_{t_0}^t$ and $T_{t_0}^t$ to represent the sample means at time $t_0 \le t < T$ for the control and treatment cohorts that joined the test at time $t_0$. 
Additionally, let $\hat{\sigma}^2_{C_{t_0}^t}$ and $\hat{\sigma}^2_{T_{t_0}^t}$ be the sample mean variances of $C_{t_0}^t$ and $T_{t_0}^t$, respectively.
Following~\cite{sadeghi2022Ms, hohnhold2015CCD}, we define the user learning $\delta_{k}$ to measure the difference in treatment effect between users that had been exposed for $k$ time steps (and had some time to adapt to the treatment) with respect to those at their first day in the experiment, and the LTE as
\begin{equation}\label{eq:LTE}
    LTE=\tau + lim_{t \to +\infty}\delta_t
\end{equation} 
representing the treatment effect after the user learning stabilized, with $\tau$ representing the treatment effect on the first exposure day.

After defining the LTE, one may be tempted to evaluate a treatment solely according to its long-term effect.
However, in product experimentation the deployment decision depends on the total value generated during the entire user lifecycle, not only on the steady-state behavior, as illustrated in the bottom part of Figure~\ref{fig:metrics_comparison}.
Moreover, the LTE fails to account for the treatment's impact on the service user base itself. 
To bridge this gap, we adopt a framework based on the Expected Remaining Lifetime Value (ERLV).
We define the Remaining Lifetime Value for a specific user $u$ as the sum of the metric values generated from the current time $t^*$ until the user churns and stop using the service. 
Since $u$ may churn at any time, we model this using a survival function $S_u(t)$, which takes the value $1$ if the user is active at time $t$ and $0$ otherwise.
The ERLV is the expectation of this quantity over the user population $ERLV = \mathbb{E}_u \left[ \sum_{t=t^*}^{\infty} m_u(t, t_0) \cdot S_u(t) \right]$ where $m_u(t, t_0)$ is the observed metric value for user $u$ at time $t$, given it entered the experiment at $t_0$. 
From a decision-making perspective, the relevant quantity is the Incremental Expected Remaining Lifetime Value ($\Delta ERLV$). 
This represents the total change in value created by the treatment, accounting for both the change in user behavior with respect to the metric $m$ and the change in retention:
\begin{equation}\label{eq:dERLV}
    \begin{aligned}
        \Delta ERLV ={}& \mathbb{E}_u\!\left[\sum_{t=t^*}^{+\infty} m_u(t,t_0) S_u(t)\mid T_u=1\right] \\
        &- \mathbb{E}_u\!\left[\sum_{t=t^*}^{+\infty} m_u(t,t_0) S_u(t)\mid T_u=0\right].
    \end{aligned}
\end{equation}
This definition shifts the focus from a single point in time (the steady state after user learning) to the area under the curve of the treatment effect, aligning with traditional lifetime value frameworks \cite{AscarzaFaderHardie2017, RossetNeumannEickVatnik2003} which prioritize the total value generated over a lifecycle rather than isolated asymptotic behaviors \cite{ankargren2024whatABMeasure}.
Lastly, this formulation naturally captures cases where a treatment might boost $m_u$ in the short term but decrease $S_u$ in the long term, resulting in a possibly negative $\Delta ERLV$ despite a potentially neutral or positive LTE.

\section{Methods}\label{sec:methods}
\subsection{A unified formulation for LTE and $\Delta ERLV$}\label{sec:unified_view}
The core of our unified framework is the observation that both $LTE$ and $\Delta ERLV$ can be derived from the same staggered-cohort data by varying the handling of missing values. 
As established in Section~\ref{sec:setup}, the $LTE$ measures the steady-state behavioral change per active user, while $\Delta ERLV$ measures the cumulative business impact accounting for user participation~\cite{AscarzaFaderHardie2017, RossetNeumannEickVatnik2003, theocharous2015adRecLTV, ankargren2024whatABMeasure}.
Mathematically, we can express the two objectives as follows:
\begin{itemize}
    \item The Long Term Effect ($LTE$) is the asymptotic difference between treatment and control when considering only available (non-missing) observations ($S_u=1$). This measures the steady-state treatment effect on users assuming they keep engaging in the system. 
    \item The Incremental Lifetime Value ($\Delta ERLV$) is the cumulative difference between treatment and control weighted by the percentage of active users. This captures the combined effect of user retention and behavioral changes such as novelty or primacy effects~\cite{sadeghi2022Ms, hohnhold2015CCD, kohavi2020trustworthy, dmitriev2017dirty}.
\end{itemize} 
While traditional A/B test analysis often treats missing data as "missing at random" and ignores it, in the context of business value, churn is an informative non-random event. 
By explicitly modeling user churn, the $\Delta ERLV$ naturally weights the treatment's behavioral impact by its ability to keep the user active.
As illustrated in Figure~\ref{fig:metrics_comparison}, this distinction is critical: a treatment might have a positive or null effect on $m_u$ among survivors ($LTE \approx 0$), but if it simultaneously accelerates churn, the survival rate decrease in the $\Delta ERLV$ sum will correctly reveal a negative business outcome. 
In the following sections, we demonstrate how our multi-cohort estimator can be applied to A/B test data to recover these distinct but complementary insights.

\subsection{Efficient multi-cohort estimation}\label{sec:estimator}
Methods for long-term effect estimation under user learning first infer the learning component in short A/B tests and then use it to estimate the LTE by fitting some models and analyzing their behavior at $t\rightarrow+\infty$~\cite{hohnhold2015CCD,sadeghi2022Ms}. 
Accordingly, we first review how state-of-the-art approaches model user learning, highlighting their limitations. 
Finally, we introduce our efficient multi-cohort estimation method, which we first apply to the user learning inference problem and we then adapt for LTE and $\Delta ERLV$ estimations.

We begin by describing two representative approaches from the literature.
The CCD approach~\cite{hohnhold2015CCD} estimates the user learning as a difference between treated users exposed at the beginning of the experiment against users that just entered in the experiment, that is $\hat{\delta}_{CCD}(t) = T_{0}^t-T_{t}^t$. 
In~\cite{sadeghi2022Ms}, the user learning is instead estimated using a DiD method computed as $\hat{\delta}_{DiD}(t) = (T_{0}^t-T_{0}^0) - (C_{0}^t-C_{0}^0)$. 
Such estimates are then used to fit an exponential model $f(t) = \gamma + \alpha e^{-\beta t}$, and the LTE is obtained by summing $\gamma$ and the treatment effect in the first user exposure day.

The main limitation of such approaches lies in their inefficient use of data. In the CCD method, each cohort contributes to the estimator only once through the $T_{t}^t$ term. By contrast, the DiD estimator ignores users who enter in the experiment after it begins.
Our approach, instead, overcomes such issue by combining multiple estimates and therefore making full use of the available data. 

We observe that the user learning at time $t$ can be computed as $\hat{\delta}_0(t) = T_{0}^t-T_{t}^t$ (as shown in~\cite{hohnhold2015CCD}) but also as $\hat{\delta}_k(t) = T_{k}^{t+k}-T_{t+k}^{t+k}$ (as suggested in~\cite{sadeghi2022Ms}) as long as $t+k<T$. 
While both estimates are unbiased, combining them with a simple average to obtain a more precise quantity is inefficient due to their different variances.
We therefore aggregate such estimates using inverse-variance weighting, which assigns greater weight to more precise estimates $\hat{\delta}_k$. This yields to the multi-cohort estimator $\hat{\delta}_{MC}(t)$, which combines information across cohorts and by construction has a lower variance than any individual $\hat{\delta}_k$~\cite{cochran1954,casella2024}.
Our approach can then be summarized as follows:
\begin{enumerate}
    \item Compute the estimates $\hat{\delta}_k(t) = T_{k}^{t+k}-T_{t+k}^{t+k}$ and their variances $\hat\sigma^2_{\hat{\delta}_k}(t)=\hat{\sigma}^2_{T_{k}^{t+k}} + \hat{\sigma}^2_{T_{t+k}^{t+k}} - 2 cov(T_{k}^{t+k},T_{t+k}^{t+k})$ for $0\le t < T$ and $0\le k < T-t$.
    \item Compute the estimates weights $w_{k}(t) = \frac{1/\hat\sigma^2_{\hat{\delta}_k}(t)}{\sum_{k=0}^{T-t} 1/ \hat\sigma^2_{\hat{\delta}_k}(t)}$.
    \item Compute the weighted mean $\hat{\delta}_{MC}(t) = \sum_{k=0}^{T-t} w_{k}(t)\hat{\delta}_k(t)$ and the weighted variance $\hat\sigma^2_{\hat{\delta}_{MC}}(t) = \Biggl(\sum_{k=0}^{T-t} \frac{1}{\hat\sigma^2_{\hat{\delta}_k}(t)}\Biggr)^{-1}$ (as shown in~\cite{casella2024}).
\end{enumerate}

Such approach can also be used for LTE and $\Delta ERLV$ estimation.
To estimate the LTE, we first compute $\tau + \delta_t$ using our multi-cohort estimator applied to $T_{k}^{t+k} - C_{k}^{t+k}$, rather than $T_{k}^{t+k} - T_{t+k}^{t+k}$, ignoring missing values when computing the estimation. 
Following~\cite{sadeghi2022Ms}, we then fit an exponential model $f(t)$ to the sequence of treatment effect estimates $\hat{\tau} + \hat{\delta}_t$. 
This parametric model is finally used to extrapolate the total treatment effect beyond the observation window, that is according to Equation~\ref{eq:LTE}, $ LTE = \lim_{t \to +\infty} f(t) = \gamma$.
Estimating $\Delta ERLV$ requires instead modeling treatment and control outcomes, as well as their survival functions, separately over time. 
To this end, we again employ our multi-cohort estimator, now applied directly to $T_{k}^{t+k}$ and $C_{k}^{t+k}$ for estimating the metric values for the two cohorts. 
We then estimate $S_T(t)$ and $S_C(t)$ applying the multi-cohort method to the user presence/absence data, meaning that the metric is 1 if the user is present and 0 otherwise.
We then fit two exponential models, $f_T(t)$ and $f_C(t)$, to extrapolate the metrics trajectories for the treatment and control cohorts, respectively, and other two exponential models $f_{S_T}(t)$ and $f_{S_C}(t)$ to model user churn. 
Finally, these extrapolated quantities are combined according to Equation~\ref{eq:dERLV}, yielding $\Delta ERLV = \sum_{t=0}^{+\infty} \big(f_T(t)f_{S_T}(t) - f_C(t)f_{S_C}(t)\big)$.

\section{Experimental evaluation}
To evaluate the proposed estimators, we conduct a simulation study that reproduces realistic dynamics of user learning and churn. 
Following established practices~\cite{kohavi2020trustworthy, deng2016continuous, hohnhold2015CCD}, we use a 14-day A/A dataset consisting of user-level clicks for 41,768 users. 
This provides a neutral baseline with no pre-existing systematic differences, allowing us to inject synthetic, time-varying treatment effects and measure how accurately they are recovered.

Following the real-world setup of~\cite{hohnhold2015CCD}, we inject an exponential treatment effect $\alpha_{eff}e^{-\beta_{eff} (t-t_0)}$ to treated units. This reproduces the scenario validated by their real-world findings where a positive short-term effect decays to a negligible LTE.
To simulate staggered entry, we randomly assign each user an entry time drawn uniformly over the experimental window and shift their observed time series accordingly.
Preliminary checks on the A/A data confirm that user behavior does not differ systematically across calendar days, ensuring that this resampling does not introduce artificial biases.
In addition to learning effects, we simulate users churn in the treatment group such that the number of treated units decreases following $1 + \alpha_{churn}(e^{-\beta_{churn} (t-t_0)}-1)$. 
Control users instead follow the baseline retention observed in the A/A data. 
This design induces a scenario in which the treatment initially boosts engagement but also accelerates user churn. 
Lastly, we simulated different short A/B test durations by truncating the data after $T$ days.
% Overall, the simulated environment reflects the scenario of Figure~\ref{fig:metrics_comparison}.

\begin{table}[t!]
    \centering
    \small
    \caption{Performance comparison on 100 simulations of the evaluated methods reporting mean absolute errors for LTE and $\Delta ERLV$, and mean confidence interval size. 
    Standard deviations are in parentheses and best results are in boldface.}
    \label{tab:rw_sim_exp}
    \begin{tabular}{lccc}
        \toprule
        & $CI_{95}\ width$ & $MAE_{LTE}$ & $MAE_{\Delta ERLV}$ \\
        \midrule        
        CCD~\cite{hohnhold2015CCD}
        & 0.76 (0.08) 
        & 0.15 (0.22) 
        & 0.88 (0.83) \\
        \midrule        
        DiD~\cite{sadeghi2022Ms}
        & 1.08 (0.12) 
        & 0.16 (\textbf{0.14}) 
        & 0.88 (0.83) \\
        \midrule        
        Multi-cohort approach 
        & \textbf{0.38 (0.02)} 
        & \textbf{0.13} (0.23) 
        & \textbf{0.66 (0.50)} \\
        \bottomrule
    \end{tabular}
\end{table}

We first conducted an experiment comparing the three estimation methods. 
We ran 100 Monte-Carlo simulations, covering scenarios with and without churn. Parameters were drawn as: $T \sim U(7, 14)$, $\alpha_{eff} \sim U(0.05, 0.2)$, $\beta_{eff} \sim U(0.1, 0.5)$, $\alpha_{churn} \sim U(0, 0.1)$, and $\beta_{churn} \sim U(0.05, 0.3)$.
We then estimated the treatment effect time series using the CCD method~\cite{hohnhold2015CCD}, the DiD approach~\cite{sadeghi2022Ms} and our proposed multi-cohort estimation.
We used such results to estimate the LTE, the $\Delta ERLV$\footnote{As standard practice in such experimental evaluations~\cite{theocharous2015adRecLTV, ferrentino2016LTVMathematical,  tudoran2024understandingConsBeh}, we computed the summation up to a maximum time interval $T^*$. In our experiments $T^*=T$ days.}, and the user learning curve. We compared the LTE, the $\Delta ERLV$ with their ground truth values, and we study the user learning confidence intervals size, with results summarized in Table~\ref{tab:rw_sim_exp}.
For CCD and DiD, which do not directly estimate $\Delta ERLV$, we extended their estimate fitting the models $f_C$ and $f_T$ (Section~\ref{sec:estimator}) on the units that entered at the experiment’s start and remained consistently treated or controlled.
Table~\ref{tab:rw_sim_exp} shows that our multi-cohort estimator achieves substantially lower variance compared to CCD and DiD approaches. 
The 95\% confidence interval width of the estimates time series is reduced by 50\% relative to CCD and nearly three times relative to DiD, while the standard deviations of LTE and $\Delta ERLV$ bootstrapped estimates are also 30\% smaller. 
Moreover, our estimator is more accurate resulting in a lower mean absolute error compared to the other methods.

We then conducted another Monte-Carlo experiment with 100 simulations reproducing the scenario of Figure~\ref{fig:metrics_comparison} with: $T = 14,\ \alpha_{eff}\sim N(0.1, 0.07),\ \alpha_{churn}\sim N(0.2, 0.07),\ \beta_{eff} = \beta_{churn}= 1/3$.
At first, we analyzed the data ignoring any user learning and user churn phenomena, i.e. with standard short-term A/B tests on only the first week of data, revealing an average positive short term effect $STE = 0.32$ (std $0.01$).
We then used our multi-cohort approach to estimate both the average LTE of $0.02$ (std $0.01$) and $\Delta ERLV$ of $-0.70$ (std $0.18$).
While standard short-term analysis shows a positive immediate click increase and LTE estimates are near zero, $\Delta ERLV$ captures the treatment’s effect on user churn and highlights its overall negative effect for the business. 
This demonstrates that computing only LTE or STE could have led to the adoption of a feature that reduces long-term user value, while including $\Delta ERLV$ in the analysis prevents such a mistake.

These improvements show that combining staggered cohorts via inverse-variance weighting allows more precise and reliable inference of both LTE and $\Delta ERLV$, making it particularly suitable for short-duration experiments with user learning and churn.

\section{Conclusions}
In this paper, we have addressed two critical limitations in modern A/B testing: the high variance inherent in staggered-cohort analysis and the measurement gap between steady-state metrics and cumulative business value. 
By proposing a unified framework, we demonstrated that the Long-Term Effect (LTE) and the Incremental Expected Remaining Lifetime Value ($\Delta ERLV$) can be recovered from the same experimental data by selectively managing missing values.
Furthermore, we introduced an inverse-variance weighted multi-cohort estimator that significantly improves estimation precision. 
Our simulation study grounded in real-world A/A data confirms the practical variance reduction in the estimates.

Beyond the specific case of ad-blindness, this dual-metric approach provides a more detailed overview for other product decisions. For instance, a feature might yield $\Delta ERLV \approx 0$, indicating no immediate change in cumulative value, while maintaining a positive $LTE > 0$. 
In such cases, the framework reveals a strategic trade-off: the treatment provides a long-term benefit to power users ($LTE > 0$) at the cost of increased churn among less-engaged ones, resulting in a neutral cumulative impact ($\Delta ERLV \approx 0$).
More generally, this methodology is broadly applicable to diverse A/B testing scenarios, including search and recommendation systems where user behavior evolves over time or where experimental treatments significantly influence user retention (e.g. slow churn due to recommendation quality worsening).

Finally, our formulation bridges the gap between A/B testing and traditional Lifetime Value (LTV) frameworks~\cite{ferrentino2016LTVMathematical}. 
Unlike standard LTV models that often incorporate complex multi-year discounting over time~\cite{faderhardie2010, samuelson1937}, our $RLV$ is optimized for short- to medium-term online experiments where the discount factor is constant.
This choice is well justified in fast-moving digital products, where the high frequency of subsequent A/B tests on the same features often necessitates a focus on the steady state effect and the near term lifecycle value rather than multiple years discounted financial projections.
While traditional LTV literature focuses primarily on monetary units, our analysis operates directly on the A/B test metric $m$.
This is not a limitation, as such metrics can easily be converted into financial value using unit economics already maintained by the business.
For example, in our streaming context, a change in number of clicks can be directly translated into revenue by multiplying it by a Cost-Per-Click coefficient. 

Ultimately, this framework ensures that product decisions are informed by the total impact on the user lifecycle rather than just steady-state behavior. 
By quantifying both $LTE$ and $\Delta ERLV$, practitioners can better navigate the complex trade-offs between long-term engagement gains and user retention.

\section*{Presenter Bio}
Dario Simionato is an Applied Research Scientist at the Huawei Ireland Research Centre, where his work focuses on advancing A/B testing and causal discovery algorithms. 
His research aims to bridge the gap between theory and practice by addressing the applicability challenges inherent in deploying state-of-the-art methods to real-world environments. 
He was the recipient of the Best Paper Award at ECML PKDD 2022 and holds a Ph.D. from the University of Padua, Italy.

\bibliographystyle{ACM-Reference-Format}
\balance
\bibliography{references}

\end{document}